\documentclass{article}
\usepackage{ijcai23}

\usepackage{times}
\usepackage{soul}
\usepackage{url}
\usepackage[hidelinks]{hyperref}
\usepackage[utf8]{inputenc}
\usepackage[small]{caption}
\usepackage{graphicx}
\usepackage{amsmath}
\usepackage{amsthm}
\usepackage{booktabs}
\usepackage{tabularx}
\usepackage{algorithm}
\usepackage{algorithmic}
\usepackage[switch]{lineno}
\usepackage{amsfonts}
\usepackage{adjustbox}

\def\BibTeX{{\rm B\kern-.05em{\sc i\kern-.025em b}\kern-.08em
    T\kern-.1667em\lower.7ex\hbox{E}\kern-.125emX}}
    
\newcolumntype{C}{>{\centering\arraybackslash}X}

\title{Efficient multi-relational network representation
using primes}

\author{
Konstantinos Bougiatiotis$^{1,2}$
\and
Georgios Paliouras$^1$\and
\affiliations
$^1$Institute of Informatics and Telecommunications, National Center Scientific Research Demokritos, Athens, Greece\\
$^2$Department of Informatics and Telecommunications, National and Kapodistrian University, Athens, Greece\\
\emails
\{bogas.ko, paliourg\}@iit.demokritos.gr,
kbogas@di.uoa.gr
}

\begin{document}

\maketitle

\begin{abstract}
In this work, we propose a novel representation of complex multi-relational networks, which is compact and allows very efficient network analysis. Multi-relational networks capture complex data relationships and have a variety of applications, ranging from biomedical to financial, social, etc. As they get to be used with ever larger quantities of data, it is  crucial to find efficient ways to represent and analyse such networks. This paper introduces the concept of Prime Adjacency Matrices (PAMs), which  utilize prime numbers,  to represent the relations of the network. Due to the fundamental theorem of arithmetic, this allows for a lossless, compact representation of a complete multi-relational graph, using a single adjacency matrix. Moreover, this representation enables the fast computation of multi-hop adjacency matrices, which can be useful for a variety of downstream tasks. We illustrate the benefits of using the proposed approach through various  simple and complex network analysis tasks.
\end{abstract}

\section{Introduction}

In recent years, research on complex networks has matured and they have been the focus of study in multiple domains, such as biological, social, financial, and others~\cite{boccaletti2006complex}. This is because they allow us to model arbitrarily complex relationships between the data, thus making them very useful in real-world scenarios where complex structures arise. The observation that entities (e.g. nodes) in a complex network may be connected through multiple types of links has resulted in the study of  \textit{multi-relational} networks and their variants such as multi-layer, multi-dimensional or multi-plex networks. A good overview of these naming conventions, their definitions, and differences can be found in ~\cite{kivela2014multilayer}. Knowledge graphs~\cite{zou2020survey} are also multi-relational networks, defining relations between entities in the form of $(s, r, o)$, where $s$ and $o$ correspond to the subject and object entities, and $r$ is the relation connecting them. In this work, we will use the term multi-relational graph/network as an umbrella term to express all kinds of complex networks that can be represented through such triples.

The goal when analyzing such networks is to generate insights by aggregating the information expressed through each relation. There are many approaches to analyzing networks for different downstream tasks, such as those generating embeddings for the nodes and the relations in the graph~\cite{wang2017knowledge}, tensor decompositions~\cite{kolda2009tensor}, symbolic methodologies~\cite{ji2021survey} and more recently graph neural networks~\cite{zhou2020graph}. However, many of these approaches make use only of the direct relations between entities, without being able to capture relations that are expressed through multiple hops in the graph~\cite{sato2020survey}. In many domains~\cite{edwards2021explainable,liu2014assessment} the paths connecting entities are useful for identifying the true nature of their relationship, the role of each entity, and finally help with the task at hand. For this purpose, there is a need for a framework that will facilitate easy and fast calculations of representations that capture the rich multi-hop information of the network.

To this end, we propose the \textit{Prime Adjacency Matrix} (PAM) representation for multi-relational networks. This representation compacts, in a lossless manner, all one-hop relations of the original network in a single adjacency matrix. To do that, we take advantage of the fact that each integer can be uniquely decomposed into a collection of prime factors. By mapping each relation type  to a distinct prime, we can construct the PAM in a manner that allows us to express all the information of the original graph without loss. Then, having at our disposal one adjacency matrix for the whole graph, we can  easily calculate the powers of this matrix and generate multi-hop adjacency matrices for the graph. This process is very fast and can scale easily to large, complex networks that cover many real-world applications. 

These higher-order PAMs contain multi-hop information about the graph that can be very easily accessed; simply by looking up the values of the matrices. Hence, the rich structural information that is encapsulated in these PAMs can be used in a wide range of tasks. Specifically, we motivate multiple scenarios where this representation can be used to generate structurally-rich representations for graphs, nodes, pairs of nodes, subgraphs, etc. In this first exposure to the new representation, we design simple processes and present experimental results on tasks such as  graph classification, and relation prediction.

Specifically, the main contributions of this work are the following:
\begin{itemize}
    \item We introduce a new paradigm for representing multi-relational networks in a single adjacency matrix using primes. To the best of our knowledge, this is the first work to model the full multi-relational graph in a single adjacency matrix in a lossless fashion. 
    \item We use this compact representation for the fast calculation of multi-hop adjacency matrices for the whole complex graph, emphasizing its value for network analysis. 
    \item We prove the usefulness of the representation by conducting experiments that utilize the representation.
\end{itemize}

The rest of the paper is structured as follows: 
Section~\ref{sec:methodology} introduces the PAM framework in detail. Then we present its application on different downstream tasks in Section~\ref{sec:applications}. In Section~\ref{sec:discussion}, we comment on the current challenges of the framework and motivate possible solutions. Finally, in Section~\ref{sec:conclusions}, we summarise the main aspects of the novel method  and propose future work.\footnote{The code and related scripts can be found in the supplementary material and will also be made publicly available.}




\section{Methodology}\label{sec:methodology}
In this section, we introduce the proposed framework and highlight its main features.
\subsection{Definition}
Let us start with an unweighted, directed, multi-relational graph $G$, with $N$ nodes and $R$ unique relation types. We can represent all possible edges between the different nodes with an adjacency tensor $A$ of shape $N \times N \times R$:
\begin{equation}\label{eq:tensor_A}
        A[i,j,r]=
        \begin{cases}
            1 & \text{if $r$ connects nodes $i,j$}\\
            0 &\text{otherwise}
        \end{cases}
\end{equation}
We now associate each unique relation type $r \in R$ with a distinct prime number $p_r$, through a mapping function $\varphi$, such that: $ \forall r \in R: \varphi(r) = p_r$, where $p_r$ is prime and $p_i = p_j \iff i = j$. This mapping function is a design choice and simply allocates distinct prime numbers to each $r \in R$. At its simplest form, we would randomly order the relations and allocate the first prime to the first relation, the second prime to the second one and so forth.

With this mapping in place, we can construct the \textit{Prime Adjacency Matrix} (PAM) $P$ of shape $N \times N$ in the following form:
\begin{equation}\label{eq:PAM}
        P[i,j]=
        \begin{cases}
             \displaystyle \prod_{r:A[i,j,r]=1} p_r & if\text{ $\exists r : A[i,j,r]=1 $} \vspace{0.3em}\\ 
             \hspace{2em} $0$ &if\text{ $\forall r :$ $A[i,j,r]=0$}
        \end{cases}
\end{equation}

As we can see in \eqref{eq:PAM}, each non-zero element $P[i,j]$ is the product of the primes $p_r$ for all relations $r$ that connect node $i$ to $j$. Due to the Fundamental Theorem of Arithmetic (FTA), we can decompose each product to the original primes that constitute it (i.e. the distinct relations that connect the two nodes), thus preserving the full structure of $G$ in $P$ without any loss.

We will also define here $P_{+}$, a variant of the above matrix, which aggregates the relations between two cells through their sum instead of their product, as shown in \eqref{eq:PAM_plus}: 
\begin{equation}\label{eq:PAM_plus}
        P_{+}[i,j]=
        \begin{cases}
             \displaystyle \sum_{r:A[i,j,r]=1} p_r & if\text{ $\exists r : A[i,j,r]=1 $}\vspace{0.3em}\\
             \hspace{2em} $0$ &if\text{ $\forall r :$ $A[i,j,r]=0$}
        \end{cases}
\end{equation}
This variant is not lossless, but it is convenient for several applications, as will be explained in the following sections. As a note here, if each pair of nodes $i,j$ exhibits at most one relation between them, we can see that $P = P_{+}$, from \eqref{eq:PAM}  and \eqref{eq:PAM_plus}.

\subsection{A simple example}

\begin{figure}[htbp]

\centerline{\includegraphics{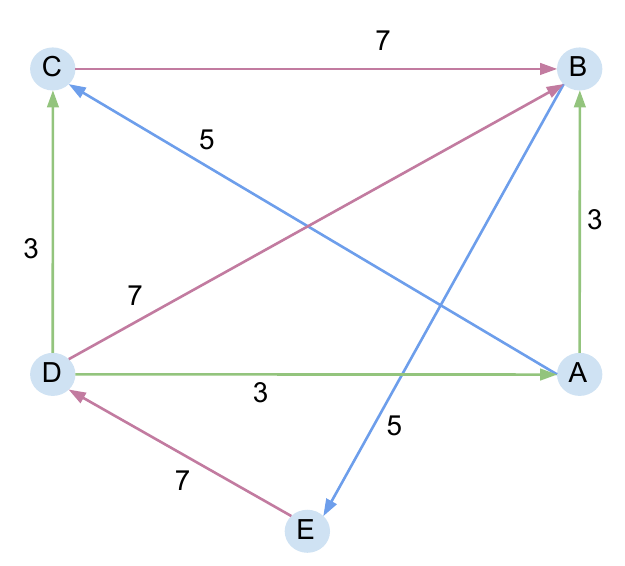}}

\caption{An example multi-relational graph with 5 nodes and 3 types of relation.}
\label{fig:PAM_rolling_example}
\end{figure}

First, let us consider the case where each pair of nodes is connected by at most one relation. In such cases, the values in $P[i,j]$ are simply the corresponding $p_r$ numbers that connect $(i,j)$. Such a graph is shown in Fig.~\ref{fig:PAM_rolling_example}, where we have 5 nodes and 3 types of relation mapped to $3$ (green), $5$ (blue) and $7$ (magenta), accordingly.

The resulting PAM would be (with node A corresponding to index 0, node B to index 1, and so forth):

\vspace{0.06cm}$P=
(\begin{smallmatrix}
0 & 3 & 5 & 0 & 0 \\
0 & 0 & 0 & 0 & 5 \\
0 & 7 & 0 & 0 & 0 \\
3 & 7 & 3 & 0 & 0 \\
0 & 0 & 0 & 7 & 0 \\
\end{smallmatrix})$.\vspace{0.06cm} 

Hence, for the edge $A\xrightarrow{3}B$ we have $P[0,1] = 3$, for $A\xrightarrow{5}C$ we have $P[0,2] = 5$ and so on, expressing all the edges in the graph.

Even in this toy graph, the compact PAM representation facilitates interesting observations. For example, we can see all the incoming/outgoing edges and their types by simply looking at the corresponding columns/rows of $P$. So, looking at $P[0,:]$ and $P[:,0]$ we see that node A has two outgoing edges (i.e. non-zero elements) of types $3$ and $5$, and one incoming edge of type $3$. Another graph property that can be easily inferred is the frequency of different relations. If we simply count the occurrences of the non-zero elements of $P$, we get the distribution of edges per relation type, which is $\{3:3, 5:2, 7:3\}$.

\subsection{Moving to multi-hop relationships}

Having a single adjacency matrix for the whole $G$ allows us to utilize tools from classical network analysis. Most importantly, we can easily obtain the powers of the adjacency matrix.  In a single-relational network,  the element $(i, j)$ of the power $k$ of an adjacency matrix, contains the number of paths of length $k$ from node $i$ to node $j$. Generalizing this property to the PAM representation, where each value in the matrix also represents a specific type of the relation, the values of $P^k[i,j]$ allow us to keep track of the relational chain linking two nodes. 

For instance, the second-order PAM for the example graph of Fig.~\ref{fig:PAM_rolling_example} will be:\newline
$P^{2}= P\times P =
(\begin{smallmatrix}
0 & 35 & 0 & 0 & 15 \\
0 & 0 & 0 & 35 & 0 \\
0 & 0 & 0 & 0 & 35 \\
0 & 30 & 15 & 0 & 35 \\
21 & 49 & 21 & 0 & 0 \\
\end{smallmatrix})$.\vspace{0.06cm}

\noindent Let us examine the values of this matrix, by starting with the node pair (A, B) for which we have $P^2[0,1] = P^2[A, B] = 35$. We can see from Fig.~\ref{fig:PAM_rolling_example}, that we can get from node A to node B in two hops only through node C, by following the directed path $A\xrightarrow{5}C\xrightarrow{7}B$. The relations $5$ and $7$ that are exhibited along this 2-hop path, are directly expressed in the value of $P^2[A,B]=35$, through its prime factors, as $35=5*7$. The same goes for the rest of the matrix $P[A,E] = 15 = 3*5$ corresponding to $A\xrightarrow{3}B\xrightarrow{5}E$, $P[E,A] = 21 = 7*3$ corresponding to $E\xrightarrow{7}B\xrightarrow{3}A$, and so on. Hence, using this representation the products in $P^k$ express the relational $k$-chains linking two nodes in the graph.

It is also important to note the case of $P^2[D, B] = 30$, which is  the sum of the two possible paths $30 = 9 + 21 = 3*3 + 3*7$, corresponding to paths $D\xrightarrow{3}A\xrightarrow{3}B$ and $D\xrightarrow{3}C\xrightarrow{7}B$ accordingly. This case shows that each cell $(i, j)$ in $P^k$ aggregates all ``path-products" of $k$-hops that lead from $i$ to $j$. This is aligned to the notion of adjacency matrix powers in classical graph theory, with the added benefit of encoding the types of relations in the value of the cell. 


Moreover, we can easily extract structural characteristics for nodes, pairs, subgraphs, and the whole graph, by looking up the PAM. For instance, we can calculate the frequency of the 2-hop paths as in the one-hop case, by simply counting the occurrences of non-zero values in $P^2$, which in this case are: $\{15:2, 21:2, 30:1, 35:4, 49:1\}$. These can be used for further analysis according to the task at hand. For example, if this was the graph of a molecule with atoms as nodes and bonds as relational edges, the observed frequent pattern $35$, corresponding to bonds denoted by $\{5, 7\}$, could be of importance for characterizing the molecule in terms of toxicity, solubility, or permeability~\cite{sharma2017toxim}

This procedure can be iterated for as many hops as we are interested in, by simply calculating the corresponding $P^k$. The values in each of these matrices will contain aggregated information regarding the relational chains of length $k$ that connect the corresponding nodes. Interesting  characteristics about these graphs and their components can then be easily extracted through simple operations.

\subsection{The generic case}

\begin{figure}[htbp]
\centerline{\includegraphics{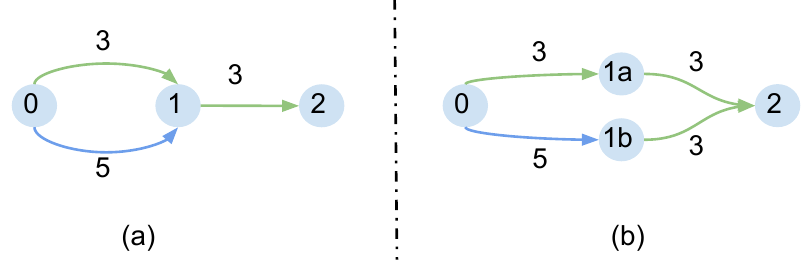}}
\caption{(a) Simple graph with multiple relations between two nodes. (b) The same graph with node 1 split into 2 nodes, 1a and 1b respectively.}
\label{fig:PAM_2hop_example}
\end{figure}

Now, let us consider the more general case where multiple relations exist between a pair of nodes. These multi-layer/plex networks, are of interest to many domains~\cite{battiston2017multilayer}. The small graph shown in Fig.~\ref{fig:PAM_2hop_example}~(a) has 3 nodes and 2 different types of relation, green and blue, mapped to the numbers $3$ and $5$ correspondingly. Moreover, the nodes $(0,1)$ are connected with relations $3$ and $5$ simultaneously.

Now let us focus on the node pair $(0,2)$. The corresponding values of the 2-hop adjacency matrix should express the sum of the 2-hop paths between these two nodes and we would expect the 2-hop value of $P^2$ to be $P^2[0,2]= 3*3 + 5*3 = 24$, aggregating the paths $0\xrightarrow{3}1\xrightarrow{3}2$ and $0\xrightarrow{5}1\xrightarrow{3}2$. To help visualise these two paths, one could split node 1 into two distinct nodes (1a) and (1b) as shown in Fig.~\ref{fig:PAM_2hop_example}~(b).

However, the PAM for this graph according to \eqref{eq:PAM} would be:\vspace{0.05cm} $P = \small(\begin{smallmatrix}
0 & 15 & 0\\
0 & 0 & 3 \\
0 & 0 & 0 \\
\end{smallmatrix}\small)$ and the corresponding $P^2 =  P \times P = \small(\begin{smallmatrix}
0 & 0 & 45\\
0 & 0 & 0 \\
0 & 0 & 0 \\
\end{smallmatrix}\small)$.\vspace{0.05cm} We can see that $P[0,1] = 3*5 = 15$,  representing the product of the relations between the nodes. This leads to $P^2[0,2] = 45 = 15*3 = P[0,1]*P[1,2]$, which is not the expected value of $24$.

The expected value can be achieved if summation was used in $P$, instead of  multiplication, i.e., $P_{+}[0,1] = 3+5 = 8 $ instead of  $P[0,1] = 3*5 = 15$. Then the corresponding PAM would be \vspace{0.1cm}$P_{+}=
\small(\begin{smallmatrix}
0 & 8 & 0\\
0 & 0 & 3 \\
0 & 0 & 0 \\
\end{smallmatrix}\small)$ and  $P_{+}^2 = P_{+} \times P_{+} = \small(\begin{smallmatrix}
0 & 0 & 24\\
0 & 0 & 0 \\
0 & 0 & 0 \\
\end{smallmatrix}\small)$\vspace{0.05cm}. 


Thus, in order to achieve a consistent representation of path aggregates, we will be using $P_{+}$ as introduced in \eqref{eq:PAM_plus}, where we simply use the sum of the primes, instead of their product. This representation is not lossless, as the sum of the primes cannot be uniquely decomposed back to the original primes. So, if  we want to represent the full graph $G$ without loss, in the $1$-hop matrix, we will need to use eq.~\eqref{eq:PAM}. When we are interested in calculating the $k$-hop PAMs, it is better to start directly with $P_{+}$ from eq.~\eqref{eq:PAM_plus}. In  the rest of the paper, when a power of PAM is used and presented as $P^{k}$, it is calculated using $P_{+}$. It is worth mentioning here that $k$-hop PAMs are lossy by design, independent of whether we use $P$ or $P_{+}$.

As a final note,  the PAM framework can easily be generalised to  heterogeneous (more than one node type) and undirected graphs. In the case of undirected graphs, we are essentially working with the upper/lower triangular PAMs. These extensions increase the applicability of the framework, which will be illustrated in the following section.

\section{Applications}\label{sec:applications}

In the following subsections, we will present some applications using  the PAM representation. All experiments were run on a Ubuntu Server with Intel Core i7 Quad-Core @ $2.30$GHz. At most, $8$ threads and $8$ GB RAM were reserved for the experiments. 

\subsection{Calculating Prime Adjacency Matrices}
To showcase the usability of the PAM representation and the simplicity of the calculations needed, we used some of the most common benchmark knowledge graphs and generated their $P^k$ matrices. Specifically, we experimented on WN18RR~\cite{dettmers2018convolutional}, YAGO3-10~\cite{DBLP:conf/cidr/MahdisoltaniBS15} FB15k-237~\cite{toutanova2015representing}, CoDEx-S~\cite{safavi2020codex},  HetioNet~\cite{himmelstein2017systematic} and ogbl-wikikg2~\cite{hu2020open}. The first three are some of the most well-known and commonly used datasets for link prediction in knowledge graphs, each with different structural characteristics. The other three datasets were selected to demonstrate the scalability of the proposed method. CoDEx-S is the smallest of three variants of the CoDEx dataset presented in ~\cite{safavi2020codex} and was selected to showcase results in a small use case. On the other hand, HetioNet and ogbl-wikikg2 are much larger, in terms of number of edges, than all other datasets used here. 
HetioNet is a biomedical network with multiple node and relation types, exhibiting also many hub nodes. 
The ogbl-wikikg2 dataset  comes  from the Open Graph Benchmark, which focuses on large-scale and challenging graphs, and was added to show the use of the approach on  a large-scale dataset.

\begin{table}[htbp]
\caption{Main characteristics of datasets and time needed to calculate PAMs up to $k=5$.}
\begin{center}
\adjustbox{max width=1\linewidth}{%
\begin{tabular}{lcccc}
\toprule
Dataset & $N$ & $R$ & $\#$ Edges & $P^5$\\
\midrule
CoDEx-S  & 2,034 & 42 & 32,888 &  0.2 sec. \\
WN18RR & 40,493 & 11 & 86,835 & 0.3 sec. \\
FB15k-237 & 14,541 & 237  & 272,115 & 39.0 sec.  \\
YAGO3-10 & 123,182 & 37 & 1,079,040  & 23.9 sec.  \\
HetioNet & 45,158 & 24 & 2,250,197  & 3.5 min.  \\
ogbl-wikikg2 & 2,500,604 & 535 & 17,137,181  & $\approx$ 40 min.   \\
\bottomrule
\end{tabular}
}
\label{tab:scalability}
\end{center}
\end{table}

The basic characteristics of the six datasets can be seen in Table~\ref{tab:scalability}, where the number of nodes, the number of unique types of relations, the total number of edges (in the training set), and the total time needed to set up PAM and calculate up to $P^5$ are presented. The datasets are sorted from small to large, based on the total number of edges in each graph. We can see that for small and medium-scale KGs the whole process takes less  than a  minute. Interestingly, the time needed to calculate $P^5$ for HetioNet is disproportionately longer  than for YAGO3-10, which is of comparable size, and this is mainly due to the structure of the dataset. Specifically, it is due to the density of the graph. It has twice the number of edges than YAGO3-10, with less than half of the nodes. Hence, it is 5 times denser, leading to denser PAMs, which in turn takes a toll on the time needed. 

It is important to note that we have not optimised the calculations of the PAMs, opting for simple sparse matrix multiplications between, based on the Compressed Sparse Row form. More efficient ways can be used to handle large-scale datasets~\cite{7013051}\footnote{For all datasets 8GB of RAM were allocated for calculating $P^5$, except for ogbl-wikikg2 which needed more than 200GB of RAM due to it's size.}. It is also worth-noting that, this procedure needs to be executed only once to calculate the needed $P^k$ and  be used for multiple downstream tasks. This means that with a few minutes of calculation we obtain the higher-order associations between graph nodes, which reveal valuable patterns. For example, using $P^5$ we can check for all the shortest paths up to 5 hops, retaining some information on the relations that are exhibited along the path. This information can, in turn, be useful  for various other tasks~\cite{ghariblou2017shortest,wu2020traffic}.


\subsection{Relation Prediction}
One task where we can utilize the complex relational patterns captured through higher-order $P^k$ is the \textit{relation prediction} task. This task consists of predicting the most probable relation that should connect two existing nodes in a graph. Essentially, we need to complete the triple (h, ?, t) where \textit{h} is the head entity and \textit{t} is the tail entity, by connecting them with a relation \textit{r}.

Our idea is that we can use the rich PAM representations of the nodes and their pairs, to construct expressive feature vectors for each pair, which can be useful for prediction, even without training a prediction model (lazy prediction). To this end, we devised a nearest neighbor scheme, where for each training sample (h, r, t), the pair (h, t) is embedded in a feature space as a sample and r is used as a label for that sample. This feature space is created by us, using the PAMs as shown next. At inference time, given a query pair ($h_q$, $t_q$), we embed it also in the same space and the missing relation is inferred via the labels/relations of its nearest neighbors. Because of the simplicity of this approach (no trainable parameters), the representation of the pairs must be rich enough to capture the semantics needed to make the correct prediction.    

To create such a representation for a given pair of nodes (h, t) we designed a simple procedure, that utilizes both information about the nodes $h$, $t$, and the paths that connect them. Specifically, the feature vector for a pair (h, t) is simply the concatenation of the representations for the head node, the tail node, and the paths that connect them. More formally, we express this as:

\begin{equation}\label{eq:rp}
    {
        R(h, t) = [Path(h, t) \| Path(t, h) \| R(h) \| R(t)]
    }
\end{equation}
where $Path(u,v)$ denotes the feature vector of the path connecting $u$ to $v$, $R(x)$ denotes the feature vector of $x$ and the symbol $\|$ denotes the concatenation of vectors. 

\begin{figure}[htbp]

\centerline{\includegraphics[width=1.1\linewidth]{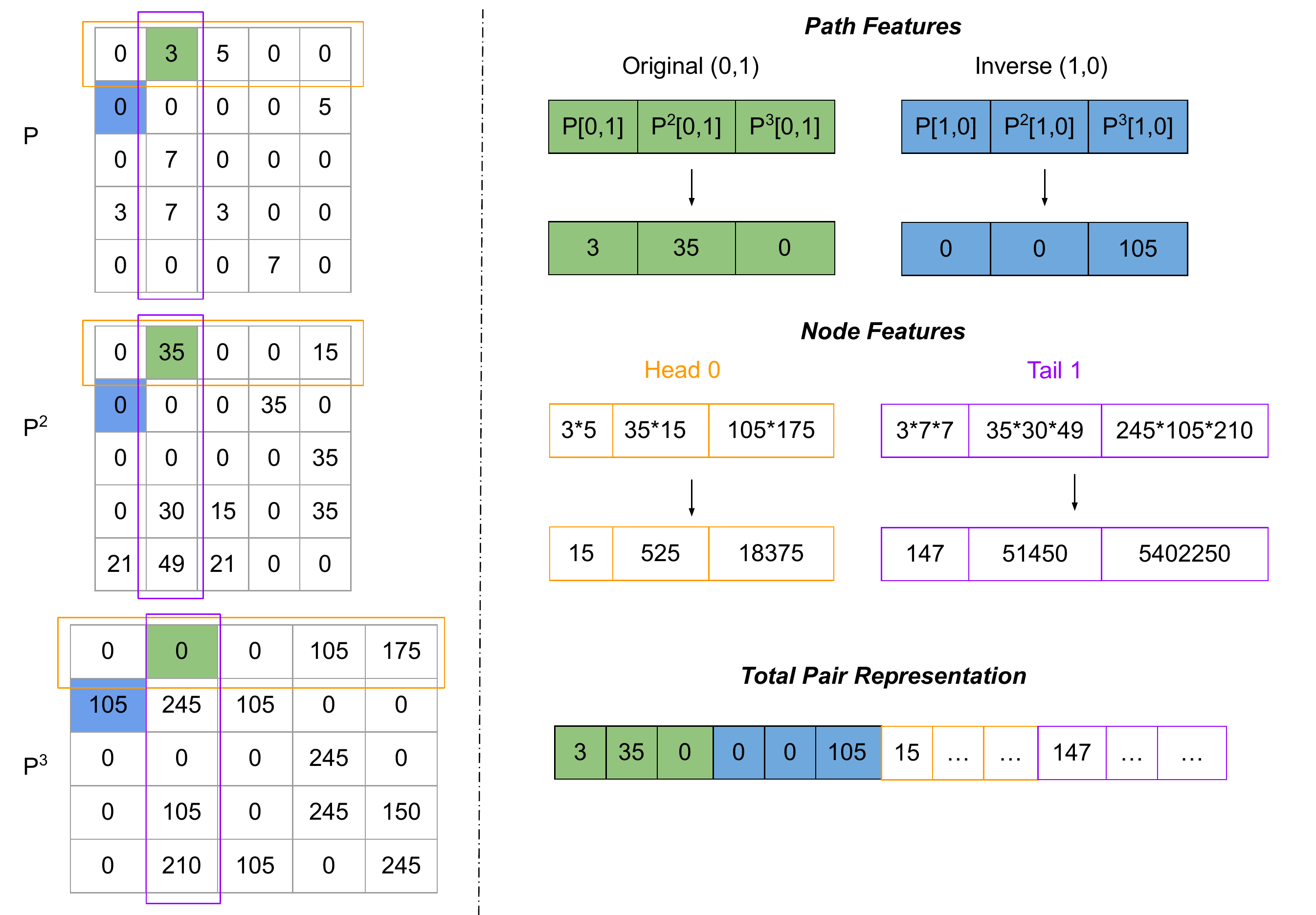}}

\caption{The construction of the feature vector of the pair (0, 1) for the graph of Fig.~\ref{fig:PAM_rolling_example}, using the PAMs up to $P^3$. On the left side, we see the PAM matrices, while on the right side the feature vector construction process. From top to bottom we have the pair features, the node features, and the final  pair representation, which is the concatenation of the individual  vectors.}
\label{fig:PAM_RelationPrediction}
\end{figure}

This procedure is highlighted in Fig.~\ref{fig:PAM_RelationPrediction} for the pair (0,1) of the small in Fig.~\ref{fig:PAM_rolling_example}. First, we create the feature vectors for the paths that connect the head to the tail and vice versa. The feature vector $Path(h, t)$ is simply a $k$-sized vector where each cell contains the corresponding value from the cell $P^k[h, t]$. That is:

$$Path(h, t) = [P[h, t], P^2[h, t], \cdots]$$.

We can see in the top-right of  Fig.~\ref{fig:PAM_RelationPrediction} that creating these path feature vectors is very easy, essentially accessing the values of the  appropriate $P^k$ matrix cells. In the example shown, the green ones correspond to the original path from $0\rightarrow1$, while the blue ones correspond to the inverse path from $1\rightarrow0$.

For the feature vectors of the head (tail) entity, we simply keep track of the products of the non-zero elements of the corresponding row (column), which essentially expresses the outgoing (incoming) relations and metapaths that the node exhibits. For the head entity, this simply is:
$$R(h) = [\prod{P[h, :]}, \prod{P^2[h, :]}, \cdots]$$ and similarly for the tail: 
$$R(t) = [\prod{P[:, t]}, \prod{P^2[:, t]}, \cdots]$$

The idea of using different representations for the head and the tail ( i.e. using the rows that represent the outgoing paths for the head, versus using the columns that represent the incoming paths for the tail), accentuates the different roles these entities play in a relational triple.

We can see this process  in the middle-right of Fig.~\ref{fig:PAM_RelationPrediction}. Here the orange feature vector corresponds to the feature vector of the head $0$ and is constructed by calculating the product of the non-zero elements of the orange-annotated rows of the matrices on the left. The same process is followed for the tail of the pair, but now using the purple-annotated columns to create the purple feature vector.

Finally, using the above representations for the paths and the entities themselves, we simply concatenate them to create the final feature vector for the pair as per eq.~\eqref{eq:rp}. This feature vector can optionally be one-hot encoded, because the values in PAMs, and therefore the values of  the vectors, express structural properties  which are closer to categorical features rather than numerical. In the example of Fig.~\ref{fig:PAM_RelationPrediction}, the outcome can be seen in the bottom right (without the one-hot encoding procedure). 

In order to evaluate this model, which we name \textit{PAM-knn} we will follow the experimental setup presented in~\cite{10.1145/3447548.3467247}. There, the authors experiment on this task with 6 KG datasets, but we will focus on the 3 most difficult ones which were: NELL995~\cite{DBLP:journals/corr/XiongHW17}, WN18RR~\cite{dettmers2018convolutional} and DDB14, which was created by the authors and
is based on the Disease Database\footnote{\url{http://www.diseasedatabase.com/}}, which is a medical database containing
biomedical entities and their relationships.

\begin{table}[htbp]
\caption{Statistics of the datasets on relation prediction. $E_{x}$ denotes the number of edges in the corresponding split.}
\begin{center}
\begin{tabular}{lccccc}
\toprule
\textbf{Dataset} & $N$ & $R$ & $E_{train}$ & $E_{val}$ & $E_{test}$ \\
\midrule
NELL995 & 63,917 & 198 & 137,465 & 5,000 & 5,000\\
WN18RR & 40,493 & 11 & 86,835 & 3,034 & 3,134\\
DDB14 & 9,203 & 14 & 36,561 & 4,000 & 4,000\\
\bottomrule
\end{tabular}
\label{tab:datasets_rp}
\end{center}
\end{table}

The characteristics of these datasets are summarized in Table~\ref{tab:datasets_rp}. We compare \textit{PAM-knn} to several widely used graph embedding models, namely TransE~\cite{bordes2013translating}, ComplEx~\cite{trouillon2016complex}, DistMult~\cite{yang2014embedding}, RotatE~\cite{sun2019rotate} and finally DRUM~\cite{sadeghian2019drum}, a model which focuses on relational paths to
make a prediction. For a given entity pair $(h, t)$ in the test set, we rank the ground-truth relation type $r$ against all other candidate relation types according to each model. We use \textit{MRR} (mean reciprocal rank) and \textit{Hit@3} (hit ratio with cut-off values of 3) as evaluation metrics, as in the original work. The performance of the competing models is reported as found in the article. Details on hyperparameters for \textit{PAM-knn} and the experimental setup can be found in the supplementary material.

\begin{table}[htbp]
\caption{Results of relation prediction on all datasets. The best results are highlighted in bold, and the best results of the competing models are highlighted with underlines.}
\adjustbox{max width=1\linewidth}{%
\begin{tabular}{l|rr|rr|rr}
\toprule
 &\multicolumn{2}{c|}{NELL995} &\multicolumn{2}{c|}{WN18RR} &\multicolumn{2}{c}{DDB14}  \\
 & MRR & H@3 & MRR & H@3 & MRR & H@3 \\
\midrule
TransE & 0.784 & 0.870 & \underline{0.841} & \underline{0.889} & \textbf{\underline{0.966}} &  0.980 \\
CompleX & 0.840 & 0.880 & 0.703 & 0.765 & 0.953  & 0.968 \\
DistMult & 0.847 & 0.891 & 0.634  & 0.720 & 0.927 & 0.961 \\
RotatE & 0.799 & 0.823 & 0.729 & 0.756 & 0.953 & 0.964 \\
DRUM & \textbf{\underline{0.854}} & \textbf{\underline{0.912}} & 0.715 & 0.740 & 0.958 & \textbf{\underline{0.987}} \\
\hline
PAM-knn & 0.740 & 0.843 & \textbf{0.852} & \textbf{0.957} & 0.915  & 0.961 \\ 
\bottomrule
\end{tabular}
}
\label{tab:rp_results}
\end{table}

The results of the experiments are presented in Table~\ref{tab:rp_results}. Once again, our goal is not to do extensive experimentation on relation prediction and propose a new state-of-the-art algorithm, but rather highlight the usefulness of the PAM framework as a simple model that heavily relies on expressive representations and performs comparably well.

We can see from Table~\ref{tab:rp_results}, that in WN18RR, \textit{PAM-knn} outperformed the competing models, while in the other 2 datasets its performance was not far from the competition. It is important to note here that \textit{PAM-knn}, has no trainable parameters and the whole procedure takes a few minutes on all datasets using a CPU, while the competing models were trained for hours using a GPU. Table~\ref{tab:rp_params} presents the number of trainable parameters of the embedding models on the smallest dataset DDB14, which are in the order of millions. Moreover, they increase further as the number of nodes in the graphs grows.

\begin{table}[htbp]
    \centering
    \caption{Number of trainable parameters of all models on DDB14.}
    
    \begin{tabular}{ccccc}
       \toprule
        TransE & ComplEx & DisMult & RotatE & PAM-knn\\
       \midrule
        3.7M & 7.4M & 3.7M & 7.4M & 0 \\
        \bottomrule
    \end{tabular}

    \label{tab:rp_params}
\end{table}

To sum up, we designed a simple model for relation prediction that relies on the expressive representations of node pairs, which can be naturally constructed using the PAMs and, as shown in the results, perform on par with many widely used graph embedding methodologies. The procedure is very fast, has no trainable parameters, and can be used as a strong baseline for relation prediction. We could devise more complex representations for the pairs or train a model using the same feature vectors in a supervised setting, but this simple approach highlights the inherent expressiveness and efficiency of the PAM framework.

\subsection{Graph Classification}
Another application in which the structural properties of the graph play an important role is graph classification. We can use the higher-order PAMs to capture complex relational patterns and describe a graph in its entirety. Such a rich and compact representation of a graph can be very useful for this task~\cite{lee2018graph}. 

\begin{figure}[htbp]

\centerline{\includegraphics[width=1\linewidth]{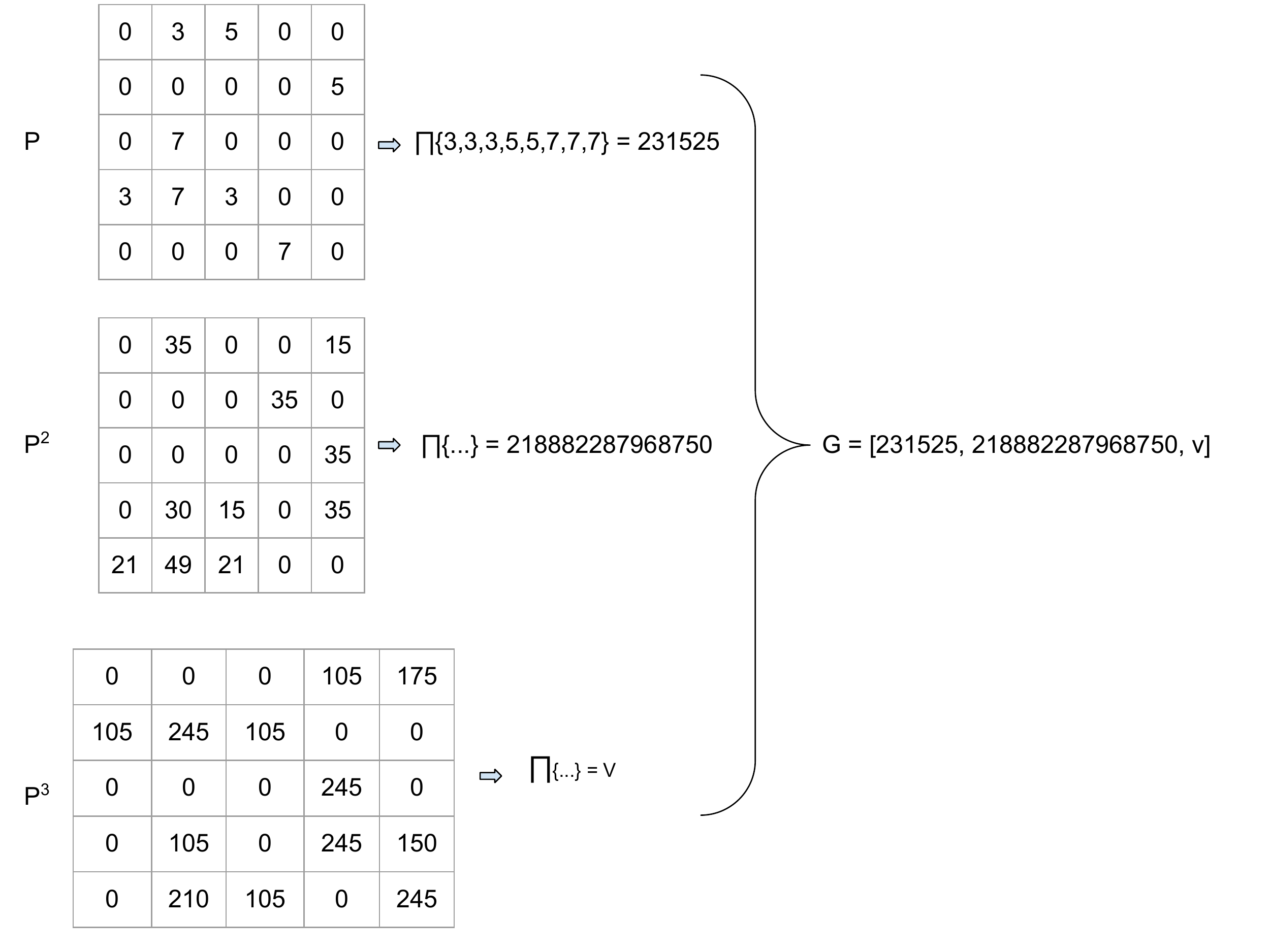}}

\caption{Graph representation for the graph of Fig.~\ref{fig:PAM_rolling_example}, using the non-zero values of PAMs up to $P^3$. For each $P^k$ we simply calculate the product of non-zero values as the $g_k$ and the final graph feature vector is their concatenation.}
\label{fig:PAM_GraphClass}
\end{figure}

To create a representation for a given multi-relational graph using PAMs we devised the procedure, highlighted in Fig.~\ref{fig:PAM_GraphClass} for the small graph of Fig.~\ref{fig:PAM_rolling_example}. First, we calculate all the PAMs  up to a pre-defined $k$ (usually up to 5 is sufficient). Then, from each matrix $P^k$ we calculate the product of its non-zero values $g_k = \displaystyle \prod_{p_{ij} > 0}{p_{ij}^k}$ as a single representative feature for this matrix. 

The intuition behind this lies in the fact that these non-zero values express the paths found at that $k$-hop. For example, as we can see in the top matrix of Fig.~\ref{fig:PAM_GraphClass}, the resulting $g_k=231525$ is the product of all the primes in the graph. By design, this number can be uniquely decomposed back to the original collection of primes (i.e. $\{3,3,3,5,5,7,7,7\}$).Therefore, it captures the information of the distribution of different relations in the graph in a single number, which acts as a ``fingerprint'' for the structure of the graph. Having generated all individual  $g_k$ values,  the final feature vector that represents the graph is simply $F(G) = [g_1, g_2, ..., g_k]$. In the example, the final feature vector has only  3 values, as we calculated only up to $P^3$, but this can be extended.

We experimented with this graph feature representation in the task of graph classification, utilizing the benchmark datasets from ~\cite{Morris+2020}. We use the multi-relational ones, which are mainly small-molecule datasets, with graphs being molecules exhibiting specific biological activities. The main characteristics of the 11 datasets used are shown in Table~\ref{tab:gc_datasets}. It is also worth noting that all the nodes have labels in these experiments (i.e. the type of the atom).  Hence, the graphs are heterogeneous.

\begin{table}[htbp]
\centering
\caption{Graph classification dataset characteristics. The reported number of nodes and edges is the average per dataset. Each dataset has 2 classes.}

\begin{tabularx}{0.95\linewidth}{lCCC}
\toprule
Dataset & $\#$ Graphs & $\#$ Nodes & $\#$ Edges \\
\midrule
AIDS & 2,000 & 15.69 & 16.20   \\
BZR$\_$MD & 306 & 21.30 & 225.06 \\
COX2$\_$MD & 303 & 26.28 & 335.12  \\
DHFR$\_$MD & 393 & 23.87 & 283.01  \\
ER$\_$MD & 446 & 21.33 & 234.85  \\
MUTAG & 188 & 17.93 & 19.79  \\
Mutagenicity & 4337 & 30.32 & 30.77 \\
PTC$\_$FM & 349 & 14.11 & 14.48 \\
PTC$\_$FR & 351 & 14.56 & 15.00 \\
PTC$\_$MM & 336 & 13.97 & 14.32 \\
PTC$\_$MR & 344 & 14.29 & 14.69 \\
\bottomrule
\end{tabularx}
\label{tab:gc_datasets}
\end{table}

We compared our approach, denoted as Power Products (PP), in terms of time and accuracy versus one of the best-performing graph kernels~\cite{DBLP:journals/corr/abs-1903-11835}, the Weisfeiler-Lehman Optimal Assignment (WL-OA) kernel~\cite{kriege2016valid}.  We used a Radial Basis Function (RBF) kernel on top of our graph feature vectors to calculate their similarity. Moreover, we created a variant which takes into account the node labels (PP-VH), using a simple Vertex Histogram (VH) kernel~\cite{sugiyama2015halting}. This will allow us to check the impact of utilizing the node information, which isn't used in our framework. As this is a classification task, we use a Support Vector Machine (SVM) (with the similarity kernel precomputed by the underlying model). We use a nested cross-validation (cv) scheme, with an outer 5-fold cv for evaluation and an inner 3-fold cv for parameter tuning. For more details on the experimental  set-up and the different hyperparameters used, please refer to the supplementary material.

The results are shown in Table~\ref{tab:gc_results}. For brevity, we report the percent change in accuracy \textit{$\Delta Acc \%$} and time \textit{$\Delta Time \%$} of PP and PP-VH over the results of WL-OA. We can see from the table that in terms of accuracy, the WL-OA kernel is better in almost all cases versus PP. However, versus PP-VH it is better only in $4/11$ cases. The impact of the addition of node information indicates that for many datasets in the small-molecule classification task, structure alone is not sufficient. For example, in the PTC$\_$FR dataset, we have an improvement of $21$ p.p. of the PP-VH over the PP variant.

\begin{table}[htbp]
\begin{center}
\caption{Results on graph classification.}

\begin{tabularx}{\linewidth}{lCCCC}
\toprule
 &\multicolumn{2}{c}{PP} &\multicolumn{2}{c}{PP-VH}  \\
 Dataset & \textit{$\Delta Acc \%$}& \textit{$\Delta Time \%$}& \textit{$\Delta Acc \%$} & \textit{$\Delta Time \%$} \\
\midrule
AIDS & -1.45 & -99.65 & +0.40  & -99.44 \\
BZR$\_$MD & +5.66 & -94.78 & +12.22 & -88.42\\
COX2$\_$MD & -20.20 & -95.79 & +0.15 & -93.26\\
DHFR$\_$MD & -16.99 & -93.78 & -48.90  & -18.50\\
ER$\_$MD & -0.38 & -92.10 & +7.45 & -89.90\\
MUTAG & -8.69 & -89.88 & +2.21  & -82.93\\
Mutagenicity & -28.53 & -99.75 & -20.23 & -95.72\\
PTC$\_$FM & -2.99 & -95.37 & -18.06 & -67.87\\
PTC$\_$FR & -10.25 & -94.16 & +10.97 & +234.68\\
PTC$\_$MM & -18.35 & -93.99 & -7.73 & -27.64\\
PTC$\_$MR & -9.83 & -95.45 & -14.39 & -72.10\\
\bottomrule

\end{tabularx}
\label{tab:gc_results}
\end{center}
\end{table}

In terms of performance over time, PP offers an improvement of more than $90\%$ over WL-OA across all datasets, thus
being orders of magnitude faster. Even when used with VH kernel, the time needed in most cases is less than half the time needed by WL-OA. Thus, using the PP-VH variant we have more than $50\%$ improvement in time, while also improving the performance in $7/11$ datasets. This indicates that PP-VH could be used as a strong baseline, that is also very fast in most cases. 

Overall, we've proposed a methodology that capitalizes on the information captured by PAMs and generates representations for graphs, that can be useful for downstream tasks. The aforementioned procedure is extremely fast and if paired with a method capable of handling node labels, can be used as a strong and efficient baseline for graph classification. Moreover, this is but one straightforward way to utilize the $P^k$ matrices to generate a representation for each graph/molecule. More complex models could be devised, as in our case we use a single feature $g_k$ to describe the whole $P^k$ matrix.

\section{Discussion}\label{sec:discussion}

Having presented the methodology in detail and showcased possible applications, we examine here some open challenges:\newline

\noindent\textbf{Order of relations}

    Due to the commutative property of multiplication, we lose the order of the relations in a path. For example, knowing that $P^{2}[A, B] = 35$ from Fig.~\ref{fig:PAM_rolling_example}, which is factorized into $35 = 5*7$, does not explicitly indicate whether the actual path is $A\xrightarrow{5}C\xrightarrow{7}B $ or $A\xrightarrow{7}C\xrightarrow{5}B$.
    
    A possible solution to this would be to keep an index of incoming/outgoing relations per node and create a heuristic to find the sequence of relations in k-hop paths, according to the characteristics of start/end nodes. For instance, knowing that node A has outgoing relations of type $3$ and $5$ only  (Fig.~\ref{fig:PAM_rolling_example}), we can be sure that the only possible one is $A\xrightarrow{5}C\xrightarrow{7}B$.\newline

\noindent\textbf{Exact decomposition of a sum of products}

    In the small graph of Fig.~\ref{fig:PAM_rolling_example} we had $P^2[D,B] = 30$ which we decomposed into two paths $ 30 = 9 + 21 = 3*3 + 3*7$. 
    In higher-order matrices, it is impractical  to keep track of  such compositions.  Although in this work we do not focus on the exact decomposition of PAM cell values, there are ways to handle such problems. For instance, one could use an Integer Linear Programming solver to find the exact decomposition of a complex path (i.e. sum of products), given the possible simple paths (i.e. products).

    As a closing note, despite the open research issues, PAMs proposed in this paper still encapsulate structural characteristics that can be very useful in a plethora of downstream tasks, as proved by the performance of the corresponding models.

\section{Conclusions}\label{sec:conclusions}
In this work, we presented the Prime Adjacency Matrix (PAM) representation for dealing with multi-relational networks. It is a compact representation that allows representing the one-hop relations of a network losslessly through a single-adjacency matrix.  This, in turn, leads to efficient ways of generating higher-order adjacency matrices, that encapsulate rich structural information. We showcased that the representations created are rich-enough to be useful in different downstream tasks. Overall, this is a new paradigm for representing multi-relational networks, which is very efficient and enables the use of tools from classical network theory. 

In the immediate future, we aim to further strengthen the methodology, by addressing the challenges discussed in Section~\ref{sec:discussion}. We are also interested in measuring  the effect of selecting different $\varphi$ mapping functions. Finally, it would be very interesting to see which ideas from single-relational analysis can be transferred to PAMs, such as spectral analysis~\cite{spielman2012spectral} or topological graph theory~\cite{gross2001topological}.
\clearpage

\bibliographystyle{named}
\bibliography{ijcai23}

\end{document}